\documentclass[review,onefignum,onetabnum]{siamart171218}
\usepackage[utf8]{inputenc}
\usepackage[T1]{fontenc}  

\usepackage{lipsum}
\usepackage{amsfonts}
\usepackage{graphicx}
\usepackage{epstopdf}
\ifpdf
  \DeclareGraphicsExtensions{.eps,.pdf,.png,.jpg}
\else
  \DeclareGraphicsExtensions{.eps}
\fi

\newsiamremark{remark}{Remark}
\newsiamremark{hypothesis}{Hypothesis}
\crefname{hypothesis}{Hypothesis}{Hypotheses}
\newsiamthm{claim}{Claim}
\newsiamthm{lemman}{Lemma}

\renewenvironment{definition}
{\par\vspace{\baselineskip}\noindent\textbf{Definition:}\begin{itshape}}
{\end{itshape}\par\vspace{\baselineskip}}

\renewenvironment{theorem}
{\par\vspace{\baselineskip}\noindent\textbf{Theorem:}\begin{itshape}}
{\end{itshape}\par\vspace{\baselineskip}}

\renewenvironment{lemma}
{\par\vspace{\baselineskip}\noindent\textbf{Lemma:}\begin{itshape}}
{\end{itshape}\par\vspace{\baselineskip}}

\newenvironment{example}
{\par\bigskip\indent\textbf{Example.} \itshape}
{\par}

\usepackage{paralist} 
\usepackage{subcaption}   
\usepackage{graphicx}
\usepackage{amsfonts}
\usepackage{amsmath}
\usepackage{tikz}
\usepackage{tikz-3dplot}

\usepackage{amsmath,amsfonts}

\def\eqref#1{equation~\ref{#1}}
\def\Eqref#1{Equation~\ref{#1}}

\def\1{\bm{1}}

\DeclareMathAlphabet{\mathsfit}{\encodingdefault}{\sfdefault}{m}{sl}
\SetMathAlphabet{\mathsfit}{bold}{\encodingdefault}{\sfdefault}{bx}{n}

\newcommand{\bfI}{{\bf I}}

\newcommand{\bfx}{{\bf x}}

\newcommand{\bfv}{{\bf v}}

\newcommand{\hf}{{\frac 12}}
\newcommand{\grad}{{\boldsymbol \nabla}}

\newcommand{\bfphi}{{\boldsymbol \phi}}
\newcommand{\bfPhi}{{\boldsymbol \Phi}}

\newcommand{\bftheta}{{\boldsymbol \theta}}

\usepackage{microtype} 
\usepackage{url}
\usepackage{algorithm}
\usepackage{algpseudocode} 
\usepackage{ifpdf}

\usepackage[normalem]{ulem}
\usepackage{xcolor}

\headers{Correction and Gradual refinement for Generative Models }{E Haber}

\title{Iterative Flow Matching - Path Correction and Gradual Refinement for Enhanced Generative Modeling}

\author{
Eldad Haber\thanks{The University of British Columbia, Vancouver, BC, Canada (\email{eldad.haber@ubc.ca})}
\and Shadab Ahamed\footnotemark[1]
\and Md. Shahriar Rahim Siddiqui\footnotemark[1]
\and Niloufar Zakariaei\footnotemark[1]
\and Moshe Eliasof\thanks{Ben-Gurion University of the Negev, Israel; University of Cambridge, United Kingdom}}

\usepackage{amsopn}

\makeatletter
\newcommand*{\addFileDependency}[1]{
  \typeout{(#1)}
  \@addtofilelist{#1}
  \IfFileExists{#1}{}{\typeout{No file #1.}}
}
\makeatother

\begin{document}
\nolinenumbers
\maketitle

\begin{abstract}
  Generative models for image generation are now commonly used for a wide variety of applications, ranging from guided image generation for entertainment to solving inverse problems. Nonetheless, training a generator is a non-trivial feat that requires fine-tuning and can lead to so-called hallucinations, that is, the generation of images that are unrealistic. In this work, we explore image generation using flow matching. We explain and demonstrate why flow matching can generate hallucinations, and propose an iterative process to improve the generation process. Our iterative process can be integrated into virtually \textbf{any} generative modeling technique, thereby enhancing the performance and robustness of image synthesis systems.
\end{abstract}



\section{Introduction}

Generative AI can be thought of as a point cloud matching problem. Given data points \( \mathbf{x}_0 \) that are sampled from a distribution \( \pi_0(\mathbf{x}) \)
and final points \( \mathbf{x}_T \) that are sampled from a distribution \( \pi_T(\mathbf{x}) \), our goal is to find a function that transforms points from \( \pi_0(\mathbf{x}) \) to \( \pi_T(\mathbf{x}) \) \cite{ruthotto2021introduction,harshvardhan2020comprehensive,wang2021deep,villani2008optimal}.
Such a transformation is often modeled as a stochastic or deterministic process that maps points from the source distribution \( \pi_0(\mathbf{x}) \) to the target distribution \( \pi_T(\mathbf{x}) \) while preserving key statistical properties \cite{albergo2023stochastic}. One approach to constructing such a transformation is through transport theory, where we seek a transport map \( {\cal T} \) such that \( \mathbf{x}_T = {\cal T}(\mathbf{x}_0) \) and the induced pushforward distribution \( {\cal T}_\# \pi_0 \) closely matches \( \pi_T \) \cite{pandey2024diffeomorphic}.

While it is straightforward to obtain such maps in low dimensions (see e.g. \cite{BB2000, HaberHoresh2014, AngenentHakerTannenbaum2003, maggi2023optimal,cuturi2013sinkhorn}), solving the problem for high dimension is difficult. A widely used alternative to finding such maps is to parameterize the transformation through a so-called generative model, such as normalizing flows, generative adversarial networks (GANs), variational autoencoders or flow matching and diffusion models. Each of these models approach the problem differently:
\begin{itemize}
\item {\bf Normalizing flows \cite{papamakarios2021normalizing,rezende2015variational}:} These models construct an invertible transformation \( f_{\bftheta} \) such that if \( \mathbf{x}_0 \sim \pi_0(\mathbf{x}) \), then \( \mathbf{x}_T = f_\theta(\mathbf{x}_0) \) follows \( \pi_T(\mathbf{x}) \). The transformation is learned by maximizing the likelihood of training data while ensuring that the Jacobian determinant of \( f_\theta \) remains tractable.

\item {\bf Generative adversarial networks (GANs) \cite{chakraborty2024ten,goodfellow2020generative, bamdad_MGAN, ricardo_baptista2025memorization}:} GANs learn to generate samples from \( \pi_T(\mathbf{x}) \) by training a generator \( G_{\bftheta} \) that maps noise \( \mathbf{z} \sim \pi_0(\mathbf{z}) \) to the data space. A discriminator network \( D_{\bfphi} \) is used to distinguish real samples from generated ones, and both networks are trained in an adversarial manner to improve the quality of the generated data.

\item {\bf Variational autoencoders (VAEs) \cite{kingma2013auto,kingma2019introduction}:} VAEs model the data distribution by learning a probabilistic latent space representation. They introduce an encoder network that maps data to a latent distribution and a decoder network that reconstructs data from latent variables. The training objective consists of maximizing a variational lower bound on the data likelihood while enforcing a structured prior on the latent space, such as a Gaussian distribution.

\item {\bf Flow matching and diffusion models \cite{lipman2022flow,lipman2024flow,albergo2023stochastic,song2019generative,ho2020denoising,dhariwal2021diffusion,ho2022classifier}:} Inspired by non-equilibrium thermodynamics, diffusion models define a stochastic process where data points undergo a forward diffusion process that gradually adds noise, transforming \( \pi_T(\mathbf{x}) \) into a simple prior (e.g., Gaussian). A neural network is then trained to approximate the reverse process, reconstructing \( \mathbf{x}_T \)
or the flow field from noisy samples, thereby generating new data points.
\end{itemize}

Each of these methods provides a different trade-off in terms of computational efficiency, sample quality, and training stability. For instance, normalizing flows provide exact likelihood estimation but are constrained by the need for invertibility. GANs generate sharp images but suffer from mode collapse and instability. Flow matching and diffusion models, while computationally expensive at inference time due to the need for numerical integration over many time steps, offer higher quality samples and stable training dynamics.

Recent advancements in generative AI seek to unify these paradigms by leveraging insights from optimal transport, score-based generative modeling, and neural ordinary differential equations (ODEs) \cite{HaberRuthotto2017}. Notably, Schr\"{o}dinger bridges \cite{de2021diffusion,shi2023diffusion} and continuous normalizing flows (CNFs) \cite{chen2018neural} reinterpret the transformation of \( \pi_0(\mathbf{x}) \) to \( \pi_T(\mathbf{x}) \) as a continuous evolution governed by a learned vector field, bridging the gap between diffusion models and traditional optimal transport techniques.

Even though various techniques are proposed to push points from $\pi_0$ to $\pi_T$, it is fairly understood that no process is perfect. In fact, each of the generative models pushes the initial distribution $\pi_0(\bfx)$ to a new distribution $\hat{\pi}_T(\bfx)$, where in general, $\hat{\pi}_T(\bfx) \not=  \pi_T(\bfx)$. This is due to issues such as sample size, model fidelity, or numerical approximation. As a result, samples from density $\hat{\pi}_T(\bfx)$ often present so-called hallucinations \cite{maleki2024ai}, which are clearly out-of-distribution samples. 
The understanding that generative models sample from the wrong distribution is important. Virtually every method attempts to overcome that by ad-hoc changes to the training process or the network architectures. Making a particular model more faithful to the target distribution $\pi_T(\bfx)$ is difficult and requires many trial and error steps.

In this work, we explore a different approach for the solution of the problem. 
We use flow matching \cite{lipman2022flow} as a way to {\bf iteratively refine} {\bf any} generative process, to successively obtain samples from densities $\hat{\pi}_1, \ldots, \hat{\pi}_k, \ldots$, where, under sufficient conditions 
\begin{eqnarray}
    \label{eq:conv}
    \mathcal{D}(\hat{\pi}_{k+1}, \pi_T) \le \mathcal{D}(\hat{\pi}_k, \pi_T) 
\end{eqnarray}
where $\mathcal{D}$ is some appropriate distance metric that is discussed. The process we propose is straightforward and easy to implement and it allows for the generation of samples that increasingly converge to samples from the target distribution.  
While our process is easy to implement, it is not achieved without a cost. Similar to virtually any gradual refinement algorithm, accuracy is obtained by increasing the computational cost. A trade-off between sample accuracy and computational complexity can be chosen based on the application and the quality of the data.

The rest of the paper is structured as follows. In \Cref{sec2}, we review the concept of flow matching (FM). In this paper, FM is used both for the generation of samples from $\hat{\pi}_1$ and as a technique to iteratively refine the samples. Using FM for the first iteration is not required, and one can use any technique. In \Cref{sec3}, we introduce the main idea of this paper - an iterative refinement to approximate the target distribution. We present two different approaches to achieve this goal and discuss their merits. In \Cref{sec4}, we propose some theoretical analysis. While it is difficult to formulate a complete theory, we show that at least for some flows, the proposed process converges. In \Cref{sec5}, we conduct a number of numerical experiments and conclude the paper.

\begin{figure}[h]
    \centering
    \includegraphics[width=1.0\linewidth]{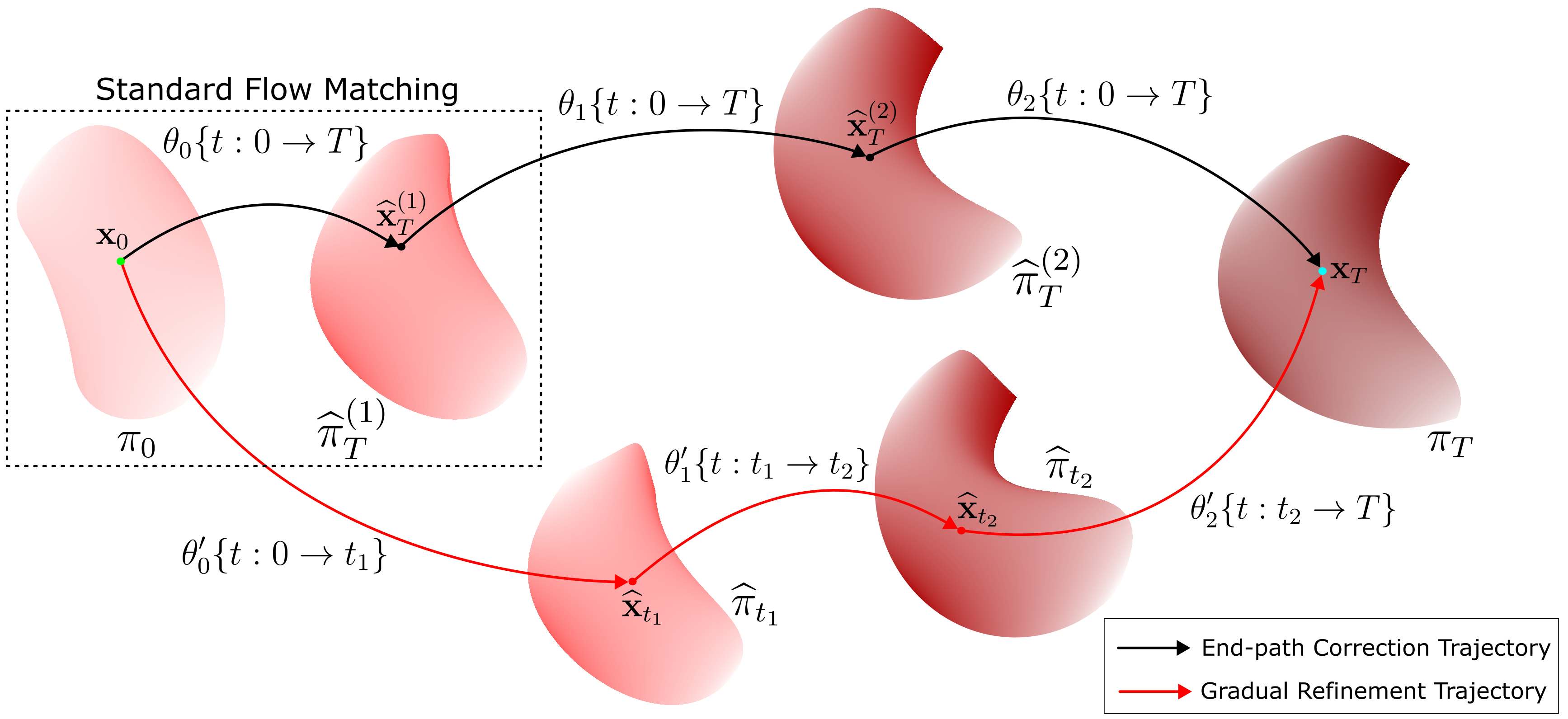}
    \caption{Schematic of the standard flow matching technique (shown inside the dashed box) vs. our proposed iterative approaches, end-path correction (black trajectory), and gradual refinement (red trajectory). $\pi_0$ and $\pi_T$ represent the source and target distributions, respectively. A trajectory between two distributions represents a learned mapping parametrized by some $\theta$. The intermediate distributions represent the pushforward distributions obtained via the integration of the ODE in \Eqref{eq:ODE} using the learned mappings $\theta$.}
    \label{fig_schematic}
\end{figure}

\section{Flow Matching}
\label{sec2}

\subsection{Flow matching - main concepts}

We now describe the concept of \emph{flow matching (FM)}, which was shown to be successful for data generation. The main idea is to generate a \emph{homotopy} $\bfx_t$ that is defined as:
\begin{equation}
    \label{eq:homo}
    \bfx_t = t \bfx_T + (T-t) \bfx_0,
\end{equation}
where $\bfx_T$ are sampled from $\pi_T(\bfx)$, $\bfx_0$ is sampled from $\pi_0(\bfx)$ and time $t \in [0,T]$. In the generative modeling literature, the construction in \Cref{eq:homo} is commonly referred to as a \textit{stochastic interpolant} \cite{albergo2023stochastic}, a particular probabilistic instance of a homotopy between distributions. We set $T=1$ following \cite{albergo2023stochastic}. The points $\bfx_t$  can be interpreted as samples from the distribution $\pi_t(\bfx)$, which are linear combinations between every point in $\pi_0$ and every point in $\pi_T$.    
For a given $\bfx_t$, the ``velocity'' or flow $\bfv$ is defined as \begin{equation}
    \label{eq:vel}
    \bfv(\bfx_t) = \bfx_T - \bfx_0.
\end{equation}
Note that the velocity field $\bfv$ is defined at points $\bfx_t$, that is, on all the trajectories that lead any point in $\pi_0$ to any point in $\pi_T$. The trajectories collide at times $t = 0$ and $t = 1$, however, in principle, they can collide or pass very close to each other even at other times. 

The main idea behind flow matching is to approximate (learn) a velocity field $\bfv(\bfx_t)$ by a function $\bfv_\bftheta(\bfx, t)$, parameterized by learnable weights $\bftheta$, by solving the stochastic optimization problem,
\begin{equation}
    \label{eq:vel_laern}
    \min_{\bftheta} \hf \int_0^T {\mathbb E}_{\bfx_t} \|\bfv_{\theta}(\bfx_t, t) - \bfv(\bfx_t) \|^2\, dt.
\end{equation}
The minimizer of \Cref{eq:vel_laern} gives rise to the best fit learned velocity which is given by the conditional expectation $ \bfv_{\bftheta^*}(\bfx, t) = \mathbb{E}[\dot{\bfx}_t \mid \bfx_t =  \bfx]$ (for notational simplicity, we henceforth omit the superscript $^*$ and denote the learned optimal parameters by $\bftheta$). In the deterministic framework, which we adopt here for simplicity, given the learned velocity model $\bfv_{\bftheta}$, one uses it at inference and integrates the ordinary differential equation (ODE),
\begin{equation}
    \label{eq:ODE}
    {\frac {d\bfx}{dt}} = \bfv_{\bftheta}(\bfx, t), \quad \quad \bfx(0) = \bfx_0 \quad {\rm where} \quad \bfx_0 \sim \pi_0(\bfx),
\end{equation}
to obtain samples from the target distribution $\pi_T(\bfx)$.

\subsection{Estimating the velocity field}

At the core of flow matching stands the solution of the optimization problem given by \Eqref{eq:vel_laern}. The problem has a very intuitive form. At time $t$, we are given some points $\bfx_t$ and a corresponding velocity $\bfv(\bfx_t)$. Our goal is to interpolate the flow field to every point $\bfx$ in space. This enables the integration of the ODE (\Eqref{eq:ODE}) for all times. In this paper, we consider two types of approximations for the flow field. In the first, $\bfv_{\bftheta}(\bfx,t)$ is parameterized by a neural network. Solving the optimization problem (\Eqref{eq:vel_laern}) is therefore done by training the network to predict $\bfv(\bfx_t)$.
This approach is particularly useful in high dimensions.

For problems in low dimensions, a simpler approach can be utilized.  We use a \textit{radial basis function} (RBF) approximation to generate the velocity field $\bfv(\bfx, t)$ given the data $\bfv(\bfx_t)$ (see \cite{buhmann2000radial} for details).
To this end, we approximately solve the linear system,
\begin{eqnarray}
    \label{eq:rbfSolve}
     \bfPhi(\bfx_t, \bfx_t) \bftheta_t = \bfv(\bfx_t)
\end{eqnarray}
using the Conjugate Gradient method, where $\bfPhi(\bfx_t, \bfx_t)$ is the kernel matrix.
We use a Gaussian exponential kernel with a tunable smoothing parameter that is a hyperparameter. Given the coefficients $\bftheta_t$, we then evaluate $\bfv(\bfx, t)$ by,
\begin{eqnarray}
    \label{eq:velx}
    \bfv(\bfx, t) =  \bfPhi(\bfx, \bfx_t) \bftheta_t = \bfPhi(\bfx, \bfx_t) (\bfPhi(\bfx_t, \bfx_t) + \beta \bfI)^{-1} \bfv(\bfx_t). 
\end{eqnarray}
where $\beta$ is a hyperparameter to avoid over-fitting and resolve inconsistencies in the data.
Putting it all together, the ODE in \Eqref{eq:ODE} can be written as,
\begin{eqnarray}
    \label{eq:ODErbf}
    \dot \bfx = \bfPhi(\bfx, \bfx_t)  (\bfPhi(\bfx_t, \bfx_t) + \beta \bfI)^{-1} \bfv(\bfx_t)
\end{eqnarray}

This approach can be easily used for problems with medium-sized matrices. For large-scale problems, several approximations can be used to invert the matrix. A common one is to cluster the data \cite{fornberg2015primer} and use the nearest neighbors, obtaining a sparse approximation.
The advantage of RBFs is their simplicity, which allows them to better explain the structure of the flow field in space.

\subsection{Numerical difficulties}

The stochastic optimization problem in \Eqref{eq:vel_laern} is a simple data-fitting problem.
We are given the flow $\bfv(\bfx)$ at points $\bfx_t$ and we desire to build an approximation $\bfv_{\bftheta}(\bfx,t)$ everywhere. Although flow matching provides a principled way to learn velocity fields, the learned field is constrained only on regions of the space–time domain where training samples exist. During numerical integration of ODE (\Cref{eq:ODE}), trajectories may drift outside this region due to discretization and approximation errors, leading to extrapolation artifacts. This issue is independent of model capacity and persists even for expressive kernel-based estimators, motivating the need for corrective or iterative schemes.


The problem is, in general, over-determined and inconsistent. Let us start at $t=0$.
At this time, a point $\bfx_0$ is matched to all points $\bfx_T$. Therefore, the flow field at $\bfx_0$ is a simple average of all the trajectories, or, since the average is linear, the flow at $\bfx_0$ points to the center of mass at $\bfx_T$. Clearly, taking a step in this direction can be off any actual trajectory. The integration of the ODE (\Eqref{eq:ODE}) requires the estimation of the velocity along the integration path. However, the velocity function is sampled over the thin lines $\bfx_t$ (tubes if some noise is added) that connect the points from $\bfx_0$ to $\bfx_T$. Since high dimensions tend to be ``empty'' (curse of dimensionality), interpolating the velocity $\bfv(\bfx)$ from points $\bfx_t$ to every point $\bfx$ in space is difficult and is prone to interpolation artifacts. Therefore, for non-degenerate source and target distributions with non-trivial support, the learned velocity $\bfv_{\bftheta}$ is, in general, time-dependent and induces trajectories that reside on curved paths. That is, the trajectories that are obtained by solving the ODE in \Eqref{eq:ODE}, in general, are not straight \cite{multiSampleFlowMatching_straighteningFlows, RectifiedFlow_liu2022flowstraightfastlearning}. On the other hand, for degenerate cases, such as Dirac-to-Dirac distribution transport, the trajectory is trivially straight and this behavior does not arise in our use-case.

Specifically, let $\bfx(t)$ be the solution of the learned ODE in \Eqref{eq:ODE}, integrated to some time $0<t<1$. Because the trajectories bend, the integrated $\bfx(t) \not= \bfx_t(t)$, that is, it deviates from the original data $\bfx_t(t)$. 
In fact, as we show in our simple example (and as have been shown in \cite{lipman2024flow} in general), $\bfx(t)$ {\em do not} reside on the line that connects two points from $\pi_0$ and $\pi_T$ and therefore represents samples from a new distribution $\widehat \pi_{t}(\bfx)$ at time $t$, that is different from  $\pi_{t}(\bfx)$.
This behavior in virtually any flow matching technique can cause significant numerical problems. The learned velocity function $\bfv_{\bftheta}(\bfx,t)$ is sampled only at points $\bfx_t(t) \sim \pi_{t}$ given by the straight  trajectories between $\bfx_0$ and $\bfx_T$.  On the other hand, the integrated point $\bfx({t})$  is drawn from the distribution $\widehat\pi_{t}(\bfx)$. During training, the network never sees data from $\widehat \pi_{t}$ but rather from $\pi_{t}$. 
It may or may not generalize well to data $\bfx(t) \sim \widehat \pi_{t}(\bfx)$. This can lead to failure of the technique to generate samples from $\bfx_T$ and generate the so-called hallucinations instead. To illustrate this behavior, we present the following example:

\begin{example}{\bf Transforming Gaussian mixtures:}\label{ex1}
    Consider the problem presented in \Cref{fig:fig1}, where we consider the movement of a Gaussian mixture made of two Gaussians, plotted in blue, to the second Gaussian mixture made from three Gaussians, plotted in red. The data of the original path $\bfx_t$ is plotted in magenta. These are straight lines that connect every sample from the blue points $\bfx_0$ to the red points $\bfx_1$. 
    We use the package that is described in \cite{lipman2024flow} and construct a neural network with roughly 50K parameters to learn the velocity $\bfv_\bftheta(\bfx, t)$, given the velocity at points $\bfx_t$ solving the stochastic optimization problem in \Cref{eq:vel_laern}. 
    However, when integrating the ODE in \Cref{eq:ODE}, one obtains the trajectories plotted in black that lead to the cyan points at $t=1$. These trajectories clearly represent the generation of data that is out-of-distribution, and therefore, many points that are generated using the learned system, yield samples that are out of the final distribution $\pi_T(\bfx)$. These samples reduce the effectiveness of the method and generate undesirable samples. 
    \begin{figure}
        \centering
        \includegraphics[width=0.45\linewidth]{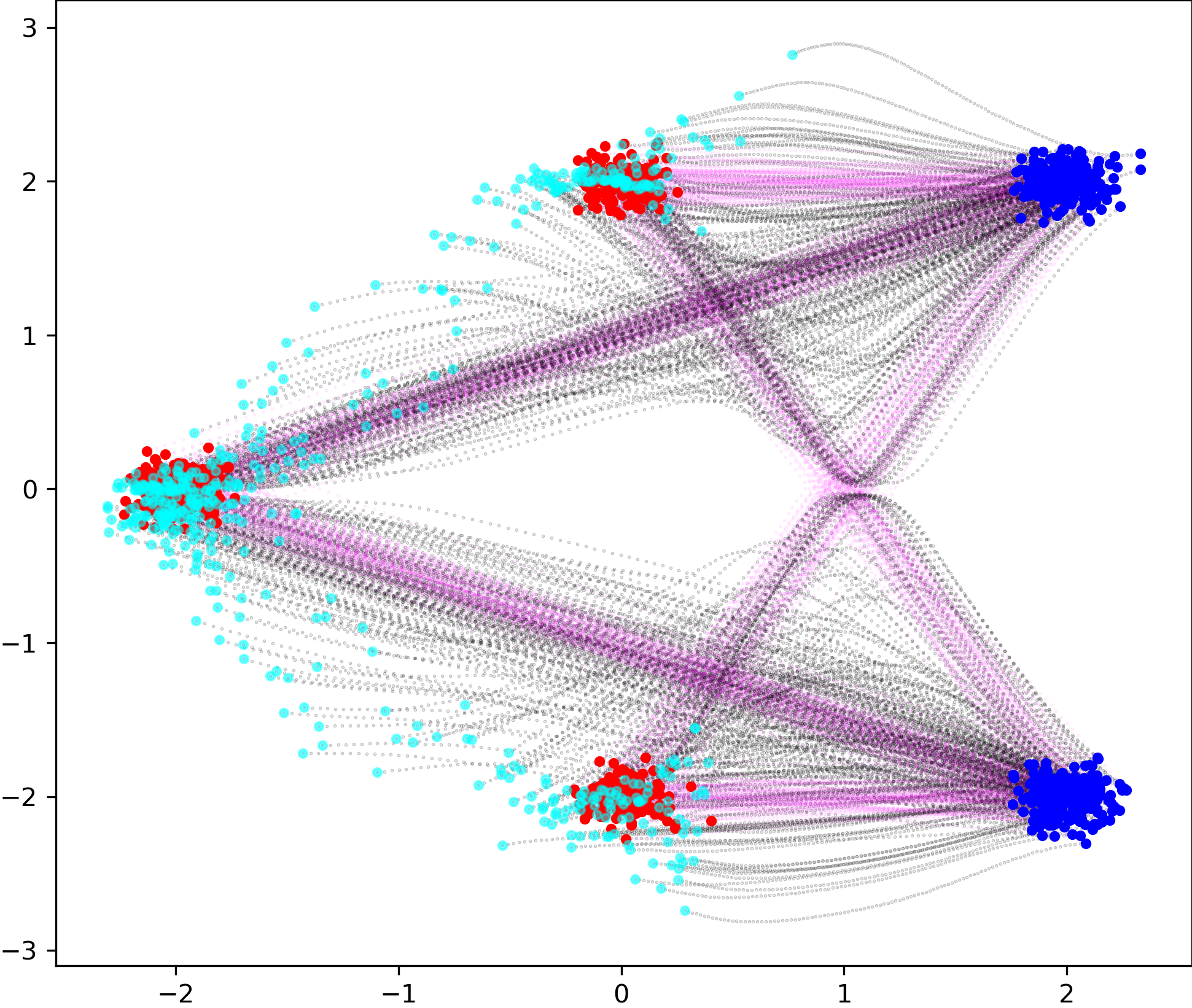}
        \caption{Using flow matching to move the distribution of blue points to the distribution of red points. The system is trained on the magenta trajectories obtained by linear combinations of points from both mixtures. The integrated data is plotted in cyan, and its trajectories are plotted in black.}
        \label{fig:fig1}
    \end{figure}
\end{example}

Given the understanding of the problem, we now propose and implement a framework that allows better sampling from the target distribution, overcoming the problems discussed above and demonstrated in Example \ref{ex1}. 

\section{Iterative Approach}
\label{sec3}
As shown in Example \ref{ex1}, the reason we obtain out-of-distribution samples from the target distribution, is that the estimated velocity $\bfv_{\bftheta}(\bfx)$, generates trajectories that do not follow the planned linear trajectory $\bfx_t$ that is computed using the linear interpolation between $\bfx_0$ to $\bfx_1$. One way to correct this problem, is to use an iterative process. The framework can be intuitively understood by anyone who has navigated from one point to another based on a predefined plan. In order to account for and correct possible trajectory drifts, rather than ``sticking'' to the original navigation plan, one has to assess their true location and update the navigation plan accordingly. Inspired by this real-life pragmatic approach, we propose two different frameworks to achieve this goal: (i) an end-path correction; and (ii) a gradual refinement approach.

\subsection{End-path correction}

In the end-path correction, we correct the final results that were obtained by using the common process described in \Eqref{eq:ODE}. This can be done iteratively to obtain, in principle, an arbitrary precision of the sampling process.

\bigskip

\noindent To this end, we consider the problem of estimating the velocity $\bfv(\bfx,t)$ given the data $\bfx_0$ and $\bfx_T$ using \Eqref{eq:vel_laern}.
Let $\bftheta_0$ be the solution of the optimization problem in \Eqref{eq:vel_laern}. Given $\bftheta_0$, we 
use the estimated velocity field and sample points $\bfx_1$ from a new distribution  $\pi_1(\bfx)$. 
The points are defined by solving the following ODE:
\begin{eqnarray}
    \label{eq:prop0}
    {\frac {d\bfx}{dt}} = \bfv_{\bftheta_0}(\bfx, t), \quad \bfx(0) = \bfx_0, \quad t \in [0,T].
\end{eqnarray}
Let $\bfx_1$ be the solution of the ODE at time $t=1$. If the flow field $\bfv_{\bftheta_0}(\bfx, t)$ is accurate, then $\bfx_1 \sim \pi_T(\bfx)$. \emph{However}, in general this is \emph{not} the case, as illustrated in \Cref{fig:fig1}. To correct this behavior, we propose to simply repeat the process but this time, using $\bfx_1$ as a starting point.
Generally, if the discrepancies $D(\pi_T, \pi_1) < D(\pi_T, \pi_0)$, then learning the correction term from $\pi_1$ to $\pi_T$ is easier than learning the solution from $\pi_0$ to $\pi_T$. The end-path approach is summarized in \Cref{alg:eppr}.

\begin{algorithm}[t]
    \caption{End-path correction for flow matching}\label{alg:eppr}
\begin{algorithmic}
\Require{$\bfx_0, \bfx_T$}
\State Set  $j=0$
\While{$\epsilon > tol$}
    \State Set $\bfx_t^{(j)} = t \bfx_T + (T-t)\bfx_j$
    \State Solve 
    \[
        \bftheta_j = \arg\min \tfrac{1}{2}\,\mathbb{E}\Bigl[\|\bfv(\bfx_t^{(j)}, t, \bftheta)
        \;-\;(\bfx_T - \bfx_j)\|^2\Bigr].
    \]
    \State Propagate $\bfx_j$, compute $\bfx_{j+1}$ by integrating
    \[
        {\frac {d\bfx}{dt}} = \bfv_{\bftheta_j}(\bfx, t) \quad \bfx(0) = \bfx_j \quad t\in [0,T]
    \]
    \State Estimate the error in samples $\bfx_{j+1}$ and stop if the error is sufficiently small.
\EndWhile
\end{algorithmic}
\end{algorithm}

The algorithm proposed here uses a number of models to approximate the path from $\bfx_0$ to $\bfx_T$. The model is actually a composite of models, each pushing forward points from the previous model toward the final distribution.
Practically, one requires a reasonable stopping criterion. To this end, a criteria that measures the similarity between two point clouds in high dimensions is needed. For problems with a small number of features, one can choose among many common criteria, for example, 
some version of the closest point \cite{wang2017survey}. For problems involving images, one can use more complex criteria, for example, Fr\'{e}chet inception distance (FID) or Inception scores \cite{chong2020effectively}. It is important to recall that for a finite sample size, the true distribution $\pi_T$ cannot be sampled exactly, and therefore, the algorithm, similar to any other algorithm, is bounded by the sample size.

We demonstrate the merit of our algorithm on the Gaussian mixture problem.
\begin{example}{\bf Gaussian mixtures with end-path correction:}
    We use 3 iterations of the proposed process. A demonstration of this algorithm for the problem of moving Gaussians is presented in \Cref{fig:alg1}.
\begin{figure}
    \centering
    \includegraphics[width=0.8\linewidth]{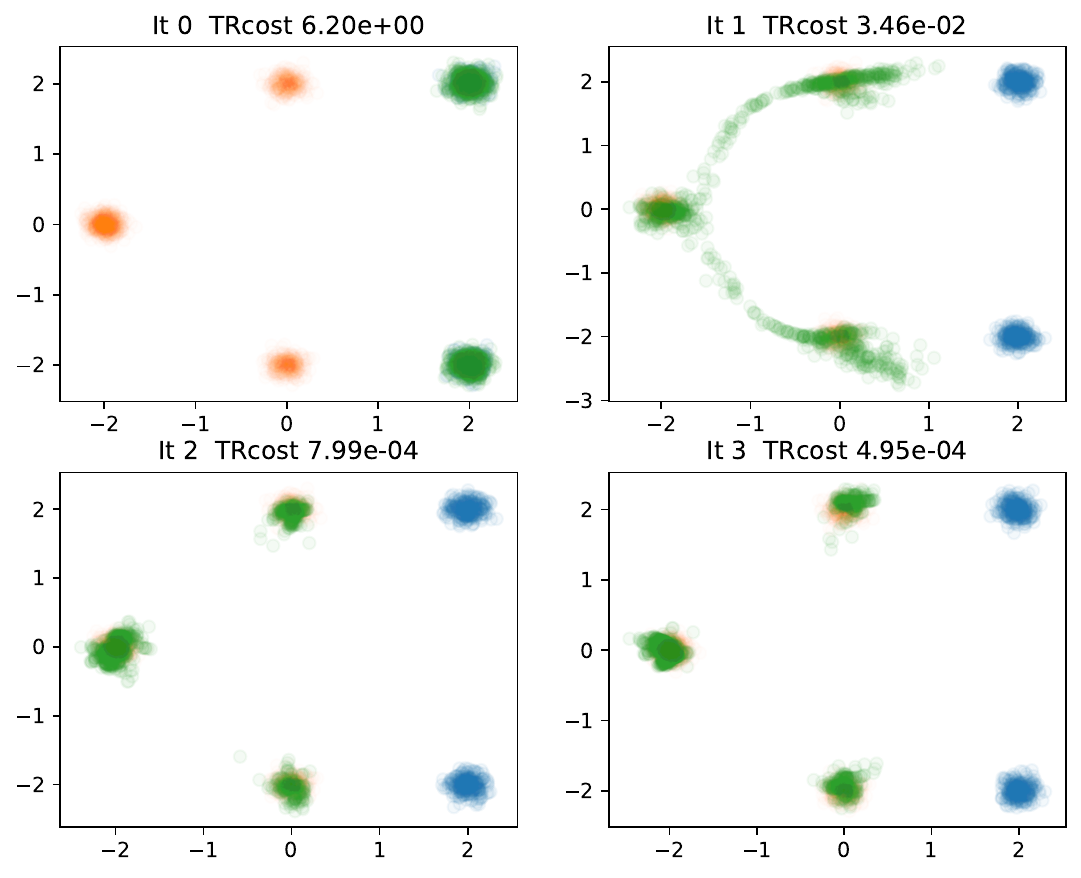}
    \caption{Using \Cref{alg:eppr} to correct the final distribution obtained from the standard flow matching optimization.}
    \label{fig:alg1}
\end{figure}
To measure the progress of our algorithm, we compute the closest point as a metric, also known as the transport cost (TRcost) \cite{memoli2004comparing}. The closest point distance between a cloud of points $\bfx_T$ and $\bfx_j$ can be written as,
\begin{eqnarray}
    \label{icp}
    \mathcal{D}(\bfx_T, \bfx_j) = {\frac 1{2N}} \left(\sum_i \min_{k} \|\bfx_T^{(k)} - \bfx_j^{(i)}\|^2 +
         \sum_i \min_{k} \|\bfx_j^{(k)} - \bfx_T^{(i)}\|^2 \right)
\end{eqnarray}
where $i$ and $k$ are indices over individual data points.

As shown in \Cref{fig:alg1}, even a single correction step can have a dramatic effect on the outcome as it reduces the distance between the point clouds in two orders of magnitudes.  
\end{example}

\subsection{Gradual refinement}

In the previous subsection, we proposed to correct the results obtained at the end of the integration path. This approach can be added as a complement to virtually any generative process.
To this end, all we need is to use any generative process to push the points $\bfx_0$ from the initial distribution $\pi_0$ to points $\bfx_1$.
However, this approach is not very efficient. It virtually integrates the path to completion only to correct it after. 
A different approach is a gradual refinement approach. Here, rather than learning the complete path, we divide the path into $n$ segments (checkpoints) at times $0 < t_1 < t_2 <, \ldots, <t_n < T$ with $\delta t_{j+1}= t_{i+1} - t_i$. We then learn and propagate the state at each segment and correct it using the learned dynamics.

Consider the first segment $[0,t_1]$. For this segment, we generate $\bfx_t$ as proposed above in \Eqref{eq:homo}, that is,
$$\bfx_t = t \bfx_T + (1-t)\bfx_0 ,\quad \quad \text{where}\quad 0 \le t \le \ t_1.$$
The network is then trained only on this interval to learn parameters $\bftheta_1$.  After the network is trained, we integrate the ODE over the interval, that is,
$$ {\frac {d\bfx_t}{dt}} = \bfv_{\bftheta_1}(\bfx_t, t) \quad t \in [0, t_1]$$
and obtain a solution $\widehat \bfx(t_1)$. As explained above, in general $\bfx_t(t_1) \not= \widehat \bfx(t_1)$. At this point, we turn to correct the homotopy path. We therefore define a new {\bf corrected} homotopy path,
$$ \bfx_t = {\frac {t-t_1}{1-t_1}} \bfx_1 + \left( 1-{\frac {t-t_1} {1-t_1}} \right)\widehat \bfx_t(t_1)$$
and a new associated velocity,
$$\bfv_t = {\frac {1}{1-t_1}} (\bfx_T - \widehat \bfx_t).$$
This corrected homotopy takes points from the actual integrated distribution at time $t_1$ towards the target $\bfx_T$ over a shorter time interval $1-t_1$.
We continue with the same strategy for every other interval. For the $j$th interval we 
have $\widehat \bfx_t(t_j)$ as the integrated solution of the ODE (\Eqref{eq:ODE}) to time $t_j$ and a corrected homotopy,
\begin{eqnarray}
    \label{eq:homoc}
    \bfx_t = {\frac {t-t_j}{1-t_j}} \bfx_T + \left( 1-{\frac {t-t_j} {1-t_j}} \right)\widehat \bfx_t(t_j) \quad {\rm and} \quad 
    \bfv_t = {\frac 1{1-t_j}} (\bfx_T - \widehat \bfx_t(t_j))
\end{eqnarray}
This idea is summarized in \Cref{alg:pr}.

\begin{algorithm}
    \caption{Gradual refinement for flow matching}\label{alg:pr}
\begin{algorithmic}
\Require{$\bfx_0, \bfx_T$}
\State Define checkpoints $0<t_1<\ldots < t_{n-1} < T $
\State Set $\widehat \bfx_t(0) = \bfx_0$
\For{$i =1, \ldots, n-1$}
    \State Choose $t_i< t < t_{i+1}$
    \State Solve the optimization problem in \Eqref{eq:vel_laern}
    \State Correct path: Solve the ODE \Eqref{eq:ODE} and set $\widehat \bfx_t(0) = \bfx_t(t_{i+1})$.
\EndFor
\end{algorithmic}
\end{algorithm}

The framework proposed above replaces the original path $\bfx_t$ with a corrected path
$\widehat \bfx_t$ that is learned iteratively. The points $\widehat \bfx_t$ corresponds to some density $\widehat \pi_t(\bfx)$.  Importantly, the points that are trained are always sampled from the correct density, and therefore, at inference, the learned velocity $\bfv_{\bftheta}(\bfx_t,t)$ is always estimated with data that belongs to distributions that have been seen in training. 
We demonstrate the effect of the method on the previous example.
\begin{example}
    {\bf Gaussian mixtures with gradual refinement:}
We solve the same problem as in Example \ref{ex1}, but this time, we use the method described in  \Cref{alg:pr}. We divide the interval $[0,1]$ into $6$ equal intervals, solve the problem for each interval, and then use its final result as a starting point for the next interval. The results for this algorithm are plotted in \Cref{fig2}. We observe that the distance between the point clouds is reduced at each iteration achieving similar accuracy as the distance obtained by the end-path approach.
   \begin{figure}[h]
        \centering
        \includegraphics[width=1.0\linewidth]{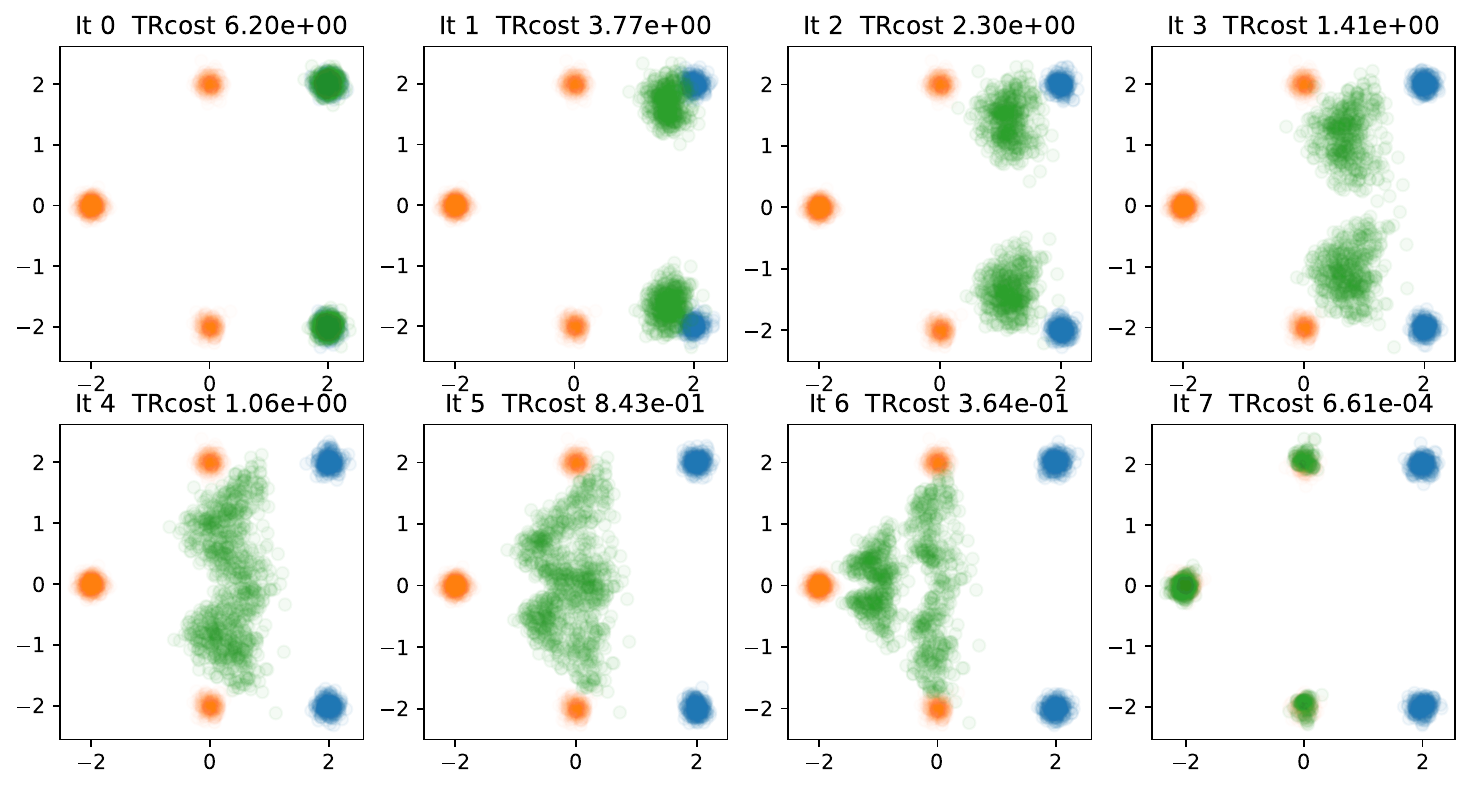}
        \caption{Using prediction-correction flow matching to move between two Gaussian mixtures. }
        \label{fig2}
    \end{figure}
When experimenting with this approach compared to the end path approach we have observed that it is less robust. One explanation is the behavior of the problem at $\bfx_0$. Note that at this time, trajectories from $\bfx_1$ to $\bfx_0$ intersect. Therefore, at this point, the velocity points to the center of mass of $\bfx_T$, which can make non-trivial trajectories. Further investigation is needed to improve this approach.
\end{example}

\section{Theoretical Properties}
\label{sec4}

At the core of our formulation stands the idea of gradually improving the learning of the transformation from $\pi_0$ to $\pi_T$. We note that in many cases, a single shot algorithm (that is, an algorithm that takes $\pi_0$ and $\pi_T$ and generates a transformation) does not yield points that faithfully represent samples from $\pi_T(\bfx)$. Our goal is to show that by using the process described in \Cref{alg:eppr,alg:pr}, we can obtain samples that better approximate the target distribution.\\

\noindent To demonstrate the merit of this idea, we rely on three key assumptions:

\begin{itemize}
\item[A1:] Given an initial distribution $\pi_0$ and a target distribution $\pi_T$, the standard FM process learns a velocity function $\bfv_{\bftheta}$ such that if we integrate points $\bfx_0 \sim \pi_0$ using the ODE (\Eqref{eq:ODE}), we obtain points $\bfx_1 \sim \pi_1$ with, $$ {\cal D}(\pi_1, \pi_T) < \gamma {\cal D}(\pi_0, \pi_T),$$ where ${\cal D}$ is an appropriate distance function and $\gamma \le c_1 < 1$. In the optimal case, $\pi_1 = \pi_T$, however, as we have discussed and demonstrated in the examples above, this is rarely the case.

\item[A2:] For some appropriate distance measure, any sufficiently smooth densities $\pi$ and $\pi_T$, and an appropriate distance metric ${\cal D}(\pi, \pi_T)$, we have that if,
$${\cal D}(\pi, \pi_T) \le c,$$ 
then there is a velocity field such that,
$$\|\bfv(\bfx_t)\|^2 = {\cal O}(c^p) \quad  p>0.$$

\item[A3:] Under some smoothness assumptions on $\bfv(\bfx)$, learning a velocity field with $\bfv(\bfx) = {\cal O}(c^p)$ is easier as $c\rightarrow 0.$
\end{itemize}

The Assumption A1 simply implies that the FM process has merit, that is, it indeed generates points that better approximate the target distribution compared to the original distribution. Recent proof of this assumption can be found in \cite{fukumizu2025flow}. 
One of the difficulties with this assumption is the level of complexity of the flow model to approximate the flow at the given points. If the flow model that we use is not sufficiently expressive, then it is impossible to generate a faithful interpolant to the flow data, and as a result, flow matching will fail. 

Assumption A2 (which is proven next) is intuitive, and it implies that given two distributions, the velocity field that maps one to the other is smaller if the distributions are more similar.
Finally, Assumption A3 connects to learning theory by stating that given a smaller $\bfv$, it is easier to learn it.

By using these assumptions and claims together, we prove that both \Cref{alg:pr,alg:eppr} converge to the target distribution.

\bigskip

We now return to A2. We assume two distributions $\pi(\bfx)$ and $\pi_T(\bfx)$ and that there is a flow $\bfv(\bfx,t)$ that maps $\pi$ to $\pi_T$. The question that we ask is, can we bound
this flow field with the distance between the distributions?

Clearly, the answer, in general, is no. We can always find flow fields that are arbitrarily large that take mass from one place and bring it to another (even to the same point). However, if we limit the type of flows, then this is easily proven. To this end, we recall the definition of the Wasserstein distance.

\begin{definition}{{\bf  Wasserstein distance.}}
Let \( \pi_0(\bfx) \) and \( \pi_T(\bfx) \) be two probability distributions. The Wasserstein distance \( W_2 \) (for \( p = 2 \)) is the minimal ``kinetic energy'' required to transport \( \pi_0(\bfx) \) to \( \pi_T(\bfx) \).

Let the flow \( \pi_t(\bfx, t) \) and a flow field \( \bfv(\bfx, t) \) be defined such that,
\begin{enumerate}
    \item \( \pi(\bfx, 0) = \pi_0(x) \) (initial distribution),
    \item \( \pi(\bfx, T) = \pi_T(\bfx) \) (final distribution),
    \item The flow \( \pi_t(\bfx, t) \) satisfies the \textbf{continuity equation}:
    \[
    \frac{\partial \pi_t}{\partial t} + \nabla \cdot (\pi_t \bfv) = 0,
    \]
    which ensures the conservation of mass during transport.
\end{enumerate}

The Wasserstein distance \( W_2 \) is then defined as,
\[
W_2(\pi_0, \pi_T)^2 = \inf_{\bfv, \pi} \int_0^T \int_X \|\bfv(\bfx, t)\|^2 \, \pi_t(\bfx, t) \, d\bfx \, dt,
\]
where the infimum is taken over all flow fields \( \bfv(\bfx, t) \) and flows \( \pi_t(\bfx, t) \) that satisfy the continuity equation and boundary conditions.
\end{definition}

An important characteristic of the optimal velocity is that it is a gradient of a scalar function \cite{BB2000}. This is summarized in the following lemma.
\begin{lemma}
    \textbf{Existence of a potential for the optimal flow \cite{BB2000}.}
    Let $\bfv(\bfx,t)$ be a smooth function that minimizes the Wasserstein distance with smooth densities $\pi_0$ and $\pi_T$, then there exists a potential $\phi(\bfx, t)$ such that,
    $$ \bfv(\bfx, t) = \grad_{\bfx}\phi(\bfx,t).$$
\end{lemma}

The following theorem is known as the Talagrand inequality
\begin{theorem}
\textbf{Talagrand Inequality} (or \textbf{Transportation Cost Inequality}) \cite{talagrand1996transportation} \\
Let \( \pi_0 \) and \( \pi_T \) be two  distributions. If \( \pi_0 \) is a log-concave distribution, then:
\[
W_2(\pi_0, \pi_T)^2 \leq 2 \, D_{\text{KL}}(\pi_T \| \pi_0),
\]
that is, the  Wasserstein distance $W_2(\pi_0, \pi_T)$ can be bounded by the Kullback-Leibler divergence of \( \pi_T \) with respect to \( \pi_0 \).
\end{theorem} 

The following is a straightforward consequence of the definition and Talagrand inequality
\begin{lemma}
    Let \( \pi_0 \) and \( \pi_T \) be two  distributions and let $\bfv(\bfx,t)$ be the velocity that yield the Wasserstein distance between $\pi_0$ and $\pi_T$ then
    \begin{eqnarray}
        {\mathbb E}_{\pi(\bfx,t)} \|\bfv(\bfx,t)\|^2 \le 2 \, D_{\text{KL}}(\pi_T \| \pi_0),
    \end{eqnarray}
\end{lemma}
where $\pi(\bfx, t)$ is a joint space-time distribution and the expectation in computed over both $\bfx$ and $t$. As the distributions $\pi_0$ and $\pi_T$ become closer, the norm of the velocity field that transports one density to the next decreases.

This somewhat trivial lemma has an important consequence that is obtained by using classic approximation theory using the radial basis functions \cite{buhmann2000radial}.
\begin{lemma}
Let $\bfv(\bfx,t) = \grad_{\bfx} \phi(\bfx,t)$ be a smooth function with
$\|\bfv(\bfx,t)\|^2 \le c$, for all $t$. Let $\phi_{\bftheta}(\bfx,t)$ be an approximation to $\phi(\bfx,t)$. 
If $\phi_{\bftheta}(\bfx,t)$ is a radial basis function  approximation, 
then for a given accuracy $\epsilon$, the number of basis functions $N$ needed to approximate the function scales as,
\begin{eqnarray}
\label{eq:bound}
N = {\cal O} \left({\frac {c}{\epsilon}} \right)^{d/s}  
\end{eqnarray}
where $d$ is the dimension of the space and $s$ is the smoothness parameter.
\end{lemma}
The implication of this lemma is simple. In principle, it implies that it is easier to 
approximate the velocity function when the distributions are closer.

Now, let us explore the meaning of this theory for \Cref{alg:eppr,alg:pr}. Both algorithms produce a sequence of points that are sampled from probabilities that are closer to the target probability. This implies that given a sufficiently complex function $\bfv_{\bftheta}(\bfx,t)$, it is possible to generate better approximations of the target density. Nonetheless, the theory strongly depends on the quality of the approximating function $\bfv_{\bftheta}(\bfx,t)$. As can be observed in \Eqref{eq:bound}, such an approximation deteriorates with the dimension of the problem. Choosing an approximation that requires a limited number of parameters and data remains challenging for this approach as well.

\section{Numerical Experiments}
\label{sec5}

In this section, we experiment with two popular datasets that demonstrate the concepts beyond the toy dataset that was explored in Example \ref{ex1}. The first is the well-known MNIST \cite{mnistlecun2009}, and the second is the more complex CIFAR-10 \cite{krizhevsky2009learning} dataset.
Similar to other successful models in the field, such as the work in \cite{dao2023flow} and Stable Diffusion \cite{rombach2022high}, we use an autoencoder that learns to represent the data in a low-dimensional latent space and performs flow matching in this latent space.
Working in latent space also allows us to easily measure
similarity differences between the obtained samples by 
using the point similarity metric \eqref{icp} which is thoroughly discussed in \cite{memoli2004comparing} as a robust metric.

\subsection{Experiments with MNIST}

One of the most common data sets to test different algorithms is the MNIST dataset, composed of 60,000 images of handwritten digits 0,...,9. Our goal is to learn a generative model that can produce similar images.

We now demonstrate that even with a very simple velocity model, that does not require a neural network, it is possible to obtain state-of-the-art results when iterating. The VAE latent space for this experiment is of size $32$, and we perform flow matching in this latent space. Reducing the problem to only 32 features allows us to use the RBF kernel approximation to the velocity field, thus using \Eqref{eq:ODErbf} for flow matching.

\begin{figure}[h]
    \centering
    \begin{tabular}{cc}
  \rotatebox{90}{Iter 0}    & \includegraphics[width=0.8\linewidth]{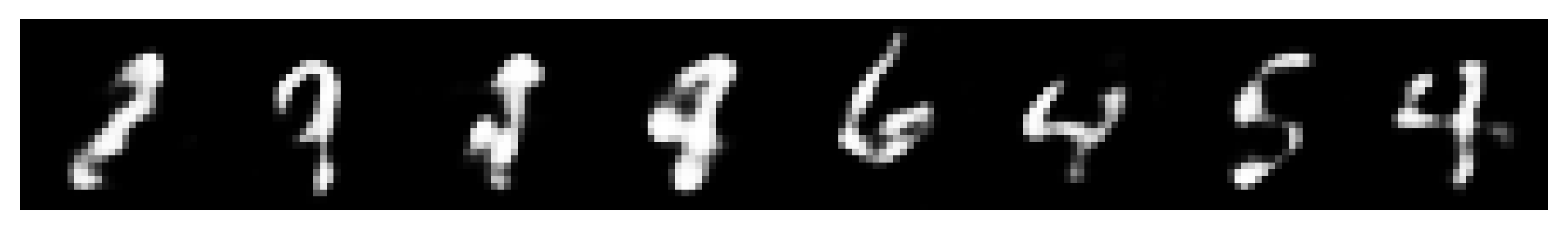} \\
   \rotatebox{90}{Iter 1}     & \includegraphics[width=0.8\linewidth]{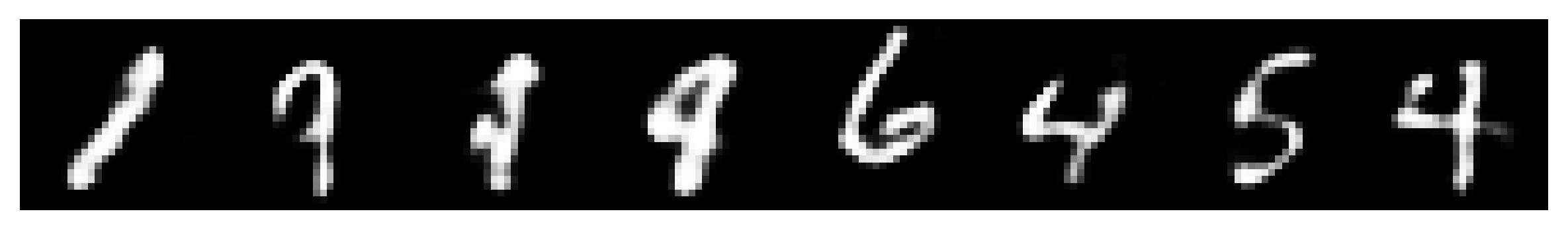} \\
   \rotatebox{90}{Iter 2}     & \includegraphics[width=0.8\linewidth]{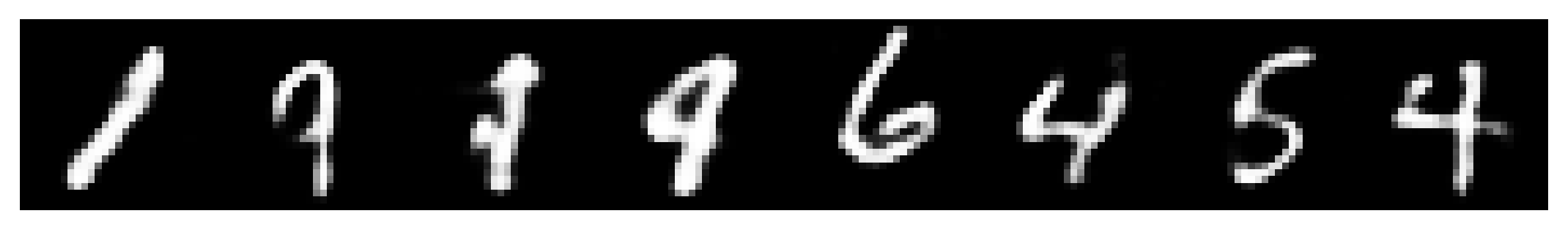} \\
   \rotatebox{90}{Iter 3}     & \includegraphics[width=0.8\linewidth]{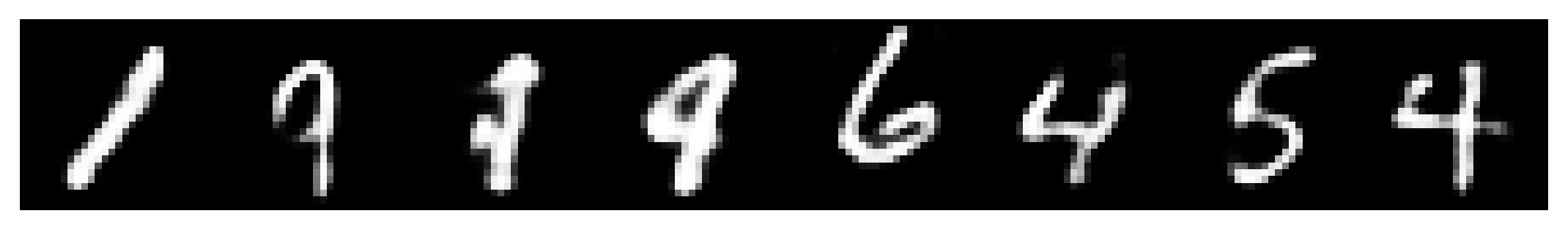} \\
   \rotatebox{90}{Iter 4}     & \includegraphics[width=0.8\linewidth]{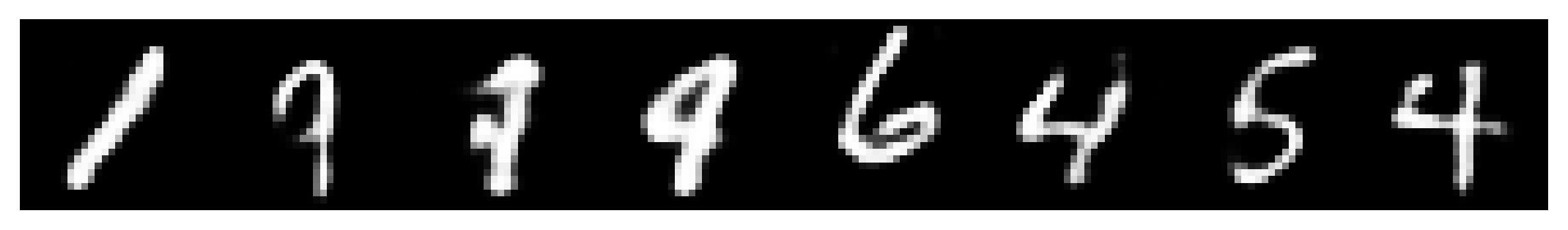} \\
   \rotatebox{90}{Iter 8}     & \includegraphics[width=0.8\linewidth]{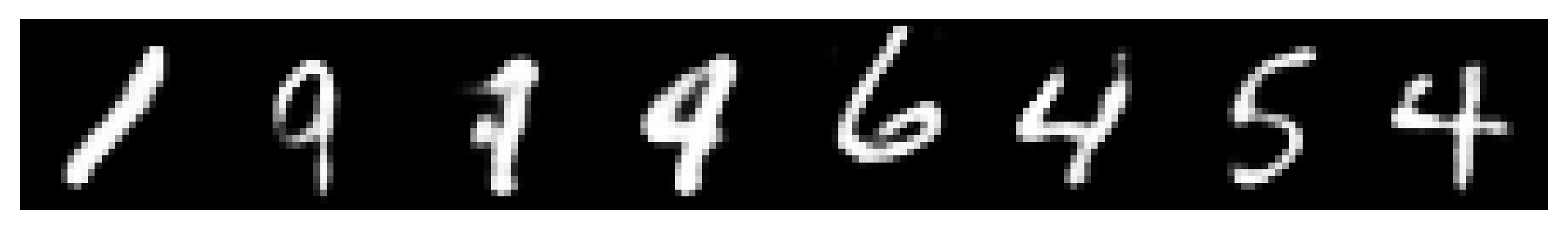} \\
   \rotatebox{90}{Iter 12}     & \includegraphics[width=0.8\linewidth]{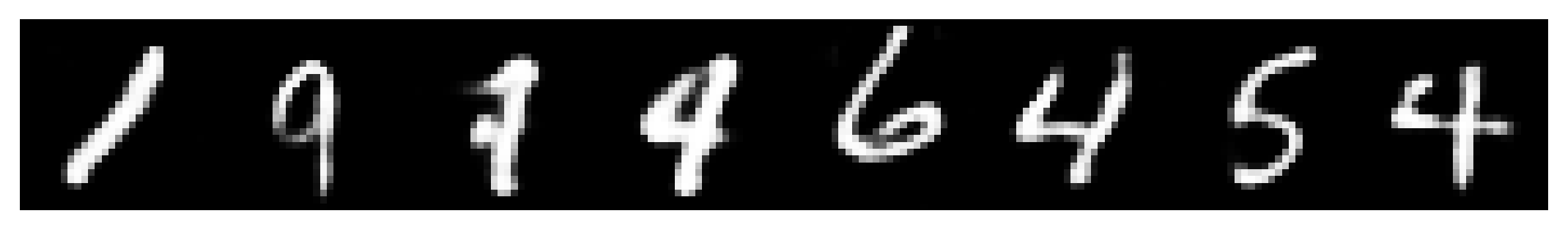} \\
   \rotatebox{90}{Iter 16}     & \includegraphics[width=0.8\linewidth]{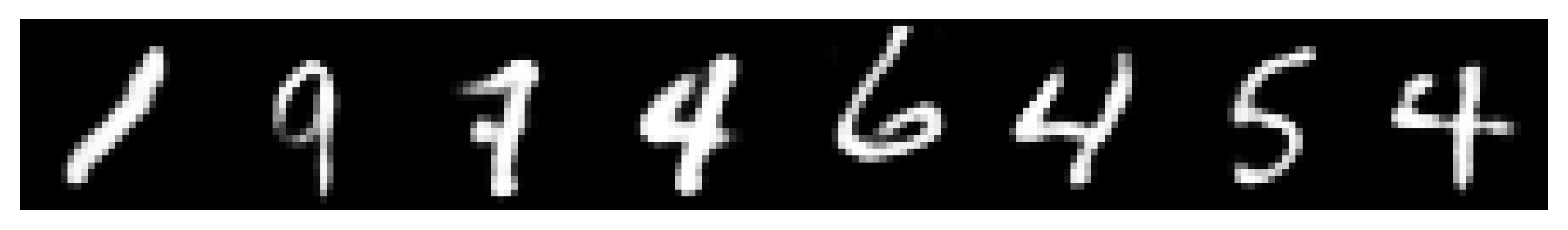} \\
   \rotatebox{90}{Iter 20}    & \includegraphics[width=0.8\linewidth]{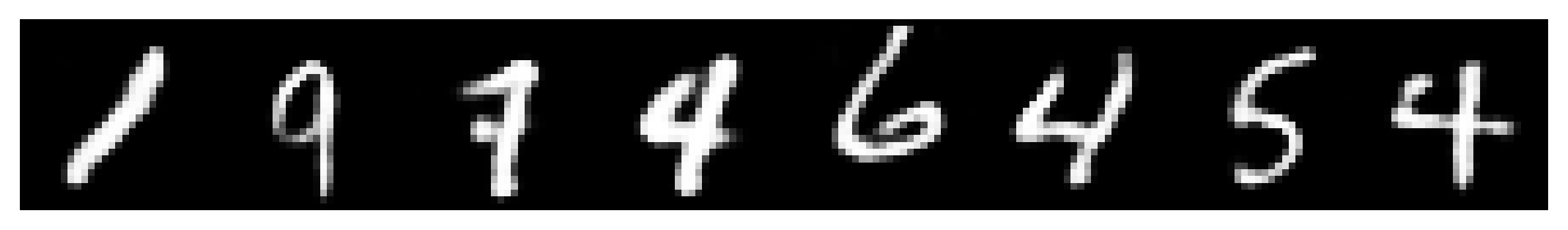} \\       
    \end{tabular}
    \caption{Generated MNIST samples from different iterations.}
    \label{fig:MNIST}
\end{figure}

\begin{figure}
    \centering
    \includegraphics[width=0.65\linewidth]{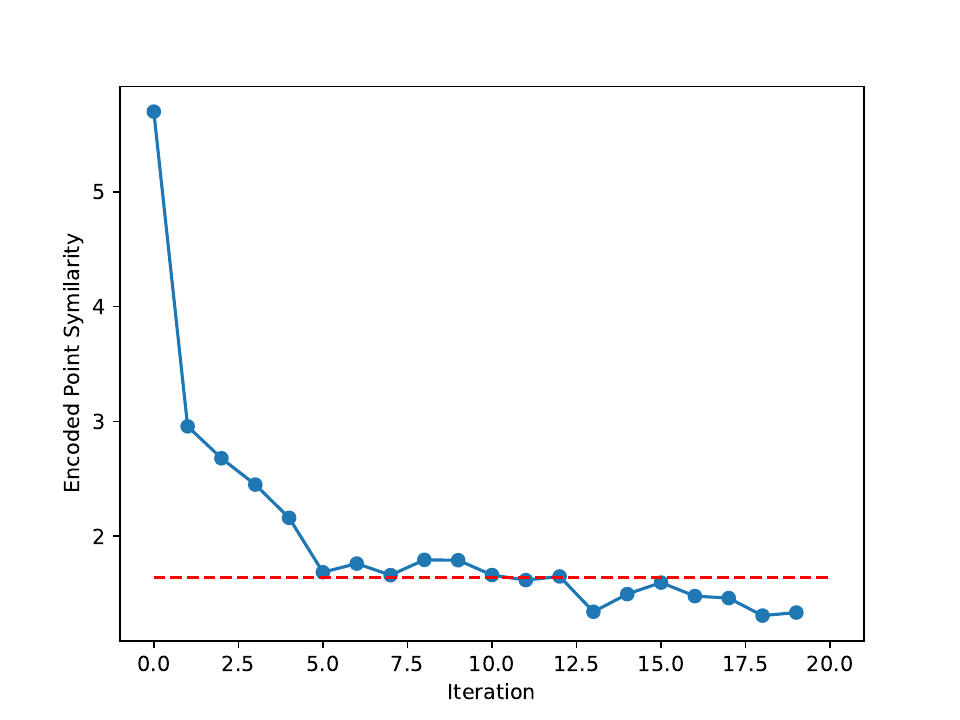}
    \caption{Similarity score for MNIST iterations. The red line is the internal self-similarity obtained by comparing 1024 samples from the data to another 1024 samples from the data.}
    \label{fig_FID_MNIST}
\end{figure}

While na\"ive kernel regression can scale as $\mathcal{O}(N^3)$, our implementation uses stochastic subset regression. In the experiments below, we initialize $\bfx_0$ to be Gaussian and $\bfx_t$ to be the linear interpolation between randomly sampled $M = 10,000$ data samples from the MNIST data and a Gaussian space. At each velocity evaluation, we sample $M$ latent pairs and solve the kernel system iteratively using conjugate gradients. We update $\bfx_0$ based on \Cref{alg:pr}. Results of the iterates, as well as the Encoded Point Similarity (EPS) score for each iteration, are presented in \Cref{fig:MNIST,fig_FID_MNIST}. It is evident that the sample quality improves as well as the EPS scores. Thus, we obtain better samples by using an iterative process compared to one-shot image generation. Finally, we compare our metrics to the internal similarity (shown by red dashed line in \Cref{fig_FID_MNIST}). The internal (self)-similarity metric is obtained by taking two random subsets of the data (in the latent space) and computing their similarity. We see that the similarity that is obtained by our method is close to the internal similarity. Increasing $M$ improves the baseline kernel approximation, but refinement addresses long-horizon integration and weakly constrained terminal regions; therefore it remains beneficial with larger $M$, though with diminishing marginal returns as the base estimator becomes more accurate.

\subsection{Experiments with CIFAR-10}

The CIFAR-10 dataset is a widely used benchmark in computer vision and machine learning research. It consists of 60,000 3-channel RGB images, each sized $32\times32$ pixels, divided into 10 distinct classes. Despite its relatively small image size, CIFAR-10's diversity in content, background, and orientation poses a significant challenge for machine learning models. 
In this experiment we have used the Unet proposed in \cite{lipman2024flow} to fit the data. However, rather than doing it in a single step we have done this in 4 different steps of learning the velocity field.
\begin{figure}
    \centering
    \begin{tabular}{cc}
    Iter 1 & Iter 2 \\
\includegraphics[width=0.45\linewidth]{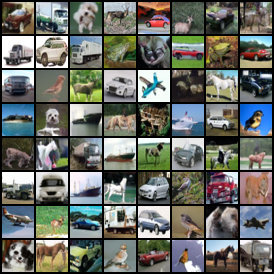} &
\includegraphics[width=0.45\linewidth]{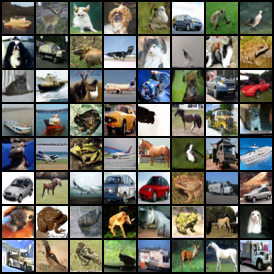} \\
Iter 3 & Iter 4 \\
\includegraphics[width=0.45\linewidth]{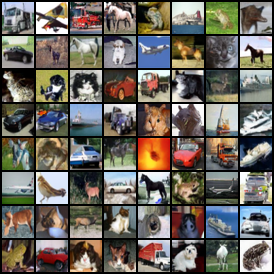} &
\includegraphics[width=0.45\linewidth]{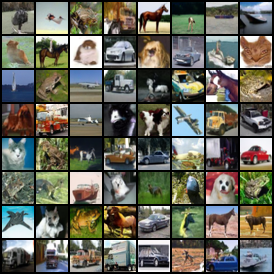}
\end{tabular}
\caption{Generated CIFAR10 samples from different iterations.}
     \label{fig:cifar}
\end{figure}

\begin{figure}
    \centering
    \includegraphics[width=0.65\linewidth]{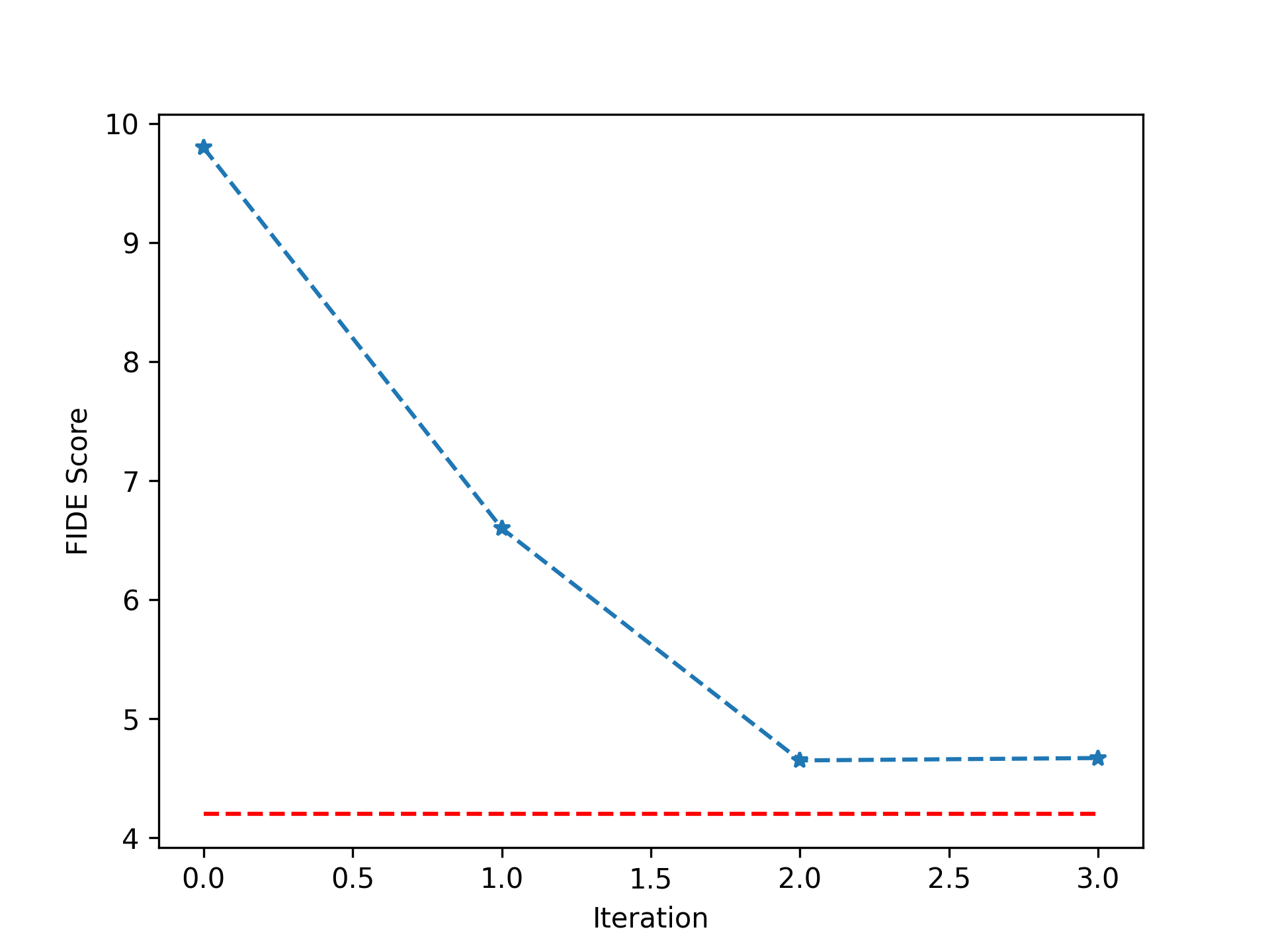}
    \caption{Similarity score for CIFAR10 iterations.}
    \label{fig_FID_CIFAR}
\end{figure}

Convergence of iterations is presented in \Cref{fig_FID_CIFAR}, where we plot the FID score  as a function of iterations.
We also plot the internal FID between points
in the data set (in red). We observe that the loss is reduced to a similar level to the internal data set loss.

\section{Discussion}

While the proposed gradual refinement strategy improves the accuracy and stability of flow matching by decomposing a difficult global transport into shorter sub-intervals, it does not eliminate all sources of numerical difficulty. In particular, diffusion- and flow-based generative models are known to exhibit near-singular behavior near one of the temporal endpoints, especially when the distribution approaches a highly concentrated or near-degenerate measure. In such regimes, even refined flows may encounter sub-intervals that remain challenging to approximate, and uniform time discretizations may be suboptimal. Our theoretical analysis does not require refinement to be uniform in time, and the refinement framework naturally accommodates non-uniform or adaptive discretizations. From a numerical perspective, this suggests that greater refinement near stiff regions (often close to one of the endpoints) may yield further gains. In practice, such regions could be identified using diagnostics derived from the learned velocity field (e.g., large velocity norms or rapid temporal variation), local integration error estimates, or distributional mismatch measures between intermediate endpoint distributions.

Another limitation of our proposed refinement strategy is that it incurs additional computational cost. As currently formulated, each refinement stage involves learning a new velocity field, and the cost of sample generation scales linearly with the number of refinement iterations. This makes the method less efficient than single-network approaches and limits its immediate applicability to large-scale, production-level generative models. Our goal in this work was not to propose a more efficient replacement for existing deployment-oriented models, but rather to address a complementary issue: the accumulation of approximation and integration errors over long time horizons in continuous-time generative models. Iterative refinement explicitly targets these errors by shortening the effective transport interval at each stage, thereby improving numerical accuracy and robustness. In this sense, the method should be viewed as a numerical correction or analysis framework rather than as a drop-in alternative to highly optimized pipelines. We note, however, that the number of required refinement stages is closely tied to how well-aligned the source and target distributions are at each step. Recent works have demonstrated that constructing more informative couplings between source and target samples, such as optimal transport paths and model-aligned couplings, can significantly straighten learned trajectories and improve sampling efficiency \cite{kong2025alignflow,lin2025beyond}. Incorporating such pairing strategies within the refinement framework has the potential to reduce the number of refinement iterations needed, and therefore the overall training and sampling cost. Exploring these combinations is a promising direction for future work.

We emphasize that the refinement strategy proposed here is complementary to existing approaches that aim to reduce endpoint stiffness through improved sample alignment or transport initialization, such as optimal transport–based matching. Recent works have shown that using optimal transport–based couplings to match noise and data samples during flow matching can improve the geometric straightness of learned trajectories and reduce endpoint stiffness \cite{kong2025alignflow,lin2025beyond}. While such methods address spatial misalignment between distributions, adaptive temporal refinement targets a distinct source of error arising from long-horizon integration and limited space–time coverage. Developing principled criteria for adaptive time discretization, and integrating them with refinement and matching strategies, represents an important direction for future work.

Our proposed iterative refinement also shares a compositional structure with greedy normalizing flow and transport-based approaches that incrementally Gaussianize a target distribution \cite{whang2021composing,meng2020gaussianization,dai2020sliced}. However, unlike these methods, our refinement is not driven by likelihood maximization or distributional simplification. Instead, it is motivated by numerical considerations in continuous-time flow matching, where long-horizon integration and limited space-time coverage can lead to extrapolation artifacts. By shortening the effective integration interval at each stage, refinement improves the local accuracy of the learned velocity field and the resulting trajectories, without altering the underlying flow matching objective.

\section{Conclusion}

Flow matching is a generative technique that can be used to match any two probability distributions. The process involves interpolating a flow field that originates from the source distribution and extends to the target distribution. Nonetheless, when interpolating the flow field from the trajectories to the whole space, interpolation artifacts can lead to flow fields that do not lead to convergence to the target distribution. To this end, we proposed an iterative refinement process that updates the trajectories. This process can, in principle, be seamlessly integrated with any generative method, providing a robust framework to enhance virtually all generative models. We demonstrate the effectiveness of this approach on a number of datasets, including one that is widely regarded as difficult and has shown that it both enhances generative performance and reduces out-of-distribution effects. We note that when redesigning the flow plan there are a number of options. Here we have explored two: end-path correction and iterative gradual refinement. While both approaches have merit, we have found experimentally that end-path correction is more robust, at least for the examples that we have considered. 

\bibliographystyle{plain}
\bibliography{biblio}

\end{document}